\documentclass[journal]{IEEEtran}
 
\usepackage{graphicx}
\usepackage{subfigure}
\usepackage{overpic}
\usepackage{amsmath}
\usepackage{flushend}
\usepackage{amssymb}
\usepackage{multirow}
\usepackage{makecell}
\usepackage{color, soul}
\sethlcolor{yellow}

\soulregister\cite7
\soulregister\citep7 
\soulregister\citet7 
\soulregister\ref7
\soulregister\pageref7
\soulregister\it7

\begin{document}

\title{Change Detection in Synthetic Aperture Radar Images Using a Dual-Domain Network}
\author{Xiaofan Qu, Feng Gao, Junyu Dong, Qian Du, Heng-Chao Li
\thanks{ This work was supported in part by the National Key Research and Development Program of China under Grant 2018AAA0100602, in part by the National Natural Science Foundation of China under Grant U1706218 and Grant 61871335, and in part by the Key Research and Development Program of Shandong Province under Grant 2019GHY112048. \emph{(Corresponding author: Feng Gao.)}

Xiaofan Qu, Feng Gao, and Junyu Dong are with the School of Information Science and Engineering, Ocean University of China, Qingdao 266100, China (e-mail: gaofeng@ouc.edu.cn).

Qian Du is with the Department of Electrical and Computer Engineering, Mississippi State University, Starkville, MS 39762 USA.

Heng-Chao Li is with the School of Information Science and Technology, Southwest Jiaotong University, Chengdu 611756, China.}
}

\markboth{}%
{Shell }

\maketitle

\begin{abstract}

Change detection from synthetic aperture radar (SAR) imagery is a critical yet challenging task. Existing methods mainly focus on feature extraction in spatial domain, and little attention has been paid to frequency domain. Furthermore, in patch-wise feature analysis, some noisy features in the marginal region may be introduced. To tackle the above two challenges, we propose a Dual-Domain Network. Specifically, we take features from the discrete cosine transform domain into consideration and the reshaped DCT coefficients are integrated into the proposed model as the frequency domain branch. Feature representations from both frequency and spatial domain are exploited to alleviate the speckle noise. In addition, we further propose a multi-region convolution module, which emphasizes the central region of each patch. The contextual information and central region features are modeled adaptively. The experimental results on three SAR datasets demonstrate the effectiveness of the proposed model. Our codes are available at \verb'https://github.com/summitgao/SAR_CD_DDNet'.

\end{abstract}


\IEEEpeerreviewmaketitle

\section{Introduction}

\IEEEPARstart{R}{cent} years have witnessed the rapid growth of synthetic aperture radar (SAR) sensors, inducing many researchers to work on SAR image segmentation, texture analysis and change detection \cite{Radke05}- \cite{Akbarizadeh12_tgrs}. As an essential task of SAR image interpretation, change detection, focusing on identifying the changed regions between multitemporal SAR images, has attracted increasing attention in remote sensing communities.

Change detection using SAR images is more challenging than optical ones due to the existence of speckle noise \cite{Davari20}. Some pioneer efforts have been made to tackle the problem of noise in multitemporal image analysis \cite{Li19}. Traditionally, the mainstream methods commonly compare multitemporal images to generate a difference image (DI) and analyze the DI to obtain the change map \cite{Celik09_grsl}. Though some pixel-wise changed information can be captured, these methods can hardly adaptively exploit the rich feature representations in the original data.

Currently, due to the popularity of deep neural networks, remote sensing image change detection methods have achieved a performance boost. Samadi et al. \cite{Samadi19} employed Deep Belief Network (DBN) for SAR image feature learning. The change detection performance has been greatly improved compared with traditional DI clustering methods. Wang et al. \cite{wang18_tgrs} presented a general end-to-end convolutional neural network (CNN) for hyperspectral image change detection. Li et al. \cite{Li19_trans} proposed a well-designed CNN to learn the spatial characteristics from raw images. In \cite{liu19_tnnls}, a local restricted CNN was proposed to enforce local similarity between neighboring pixels in change maps. 

The above-mentioned deep learning-based methods have achieved great success by exploiting deep feature representations. However, it is non-trivial to build a robust SAR change detection model, due to the following two challenges: \emph{1) Mutual reinforcement of spatial and frequency features.} Existing models are mainly based on feature extraction in the spatial domain, and little attention has been paid to the frequency domain. It has been recently shown that the compressed representation in the frequency domain is capable of suppressing the noise in the spatial domain and enriching the patterns for image understanding \cite{wu18_cvpr} \cite{xu20_cvpr}. Thereby, reinforcing spatial and frequency features in a unified framework should be considered. \emph{2) Enhancement of the central region features.} The contextual information is crucial to the performance of SAR image change detection, and hence existing methods commonly use patch-wise features for classification. However, some noisy features in the marginal region of each patch may be introduced. Therefore, how to emphasize the central region in each patch while retaining the contextual information is a tough challenge.

To tackle the above two issues, we propose a \textbf{D}ual-\textbf{D}omain \textbf{Net}work, DDNet for short, which jointly exploits the spatial and frequency features for SAR change detection task. Specifically, we first take into consideration features from the discrete cosine transform (DCT) domain. Reshaped DCT coefficients are integrated into a CNN model as a branch for inference. Therefore, feature representations from both frequency and spatial domain are exploited to alleviate the speckle noise. In addition, we further proposed a multi-region convolution (MRC) module, which emphasizes the central region of each patch. Intuitively, the contextual information and central region features are adaptively modeled efficiently.

The contributions of this letter can be summarized as follows: 

\begin{itemize}
	
\item To the best of our knowledge, we are the first to introduce features in the DCT domain to solve the SAR image change detection problem. Features from both frequency and spatial domain are exploited. Therefore, the speckle noise is suppressed efficiently.

\item We propose a MRC module, which jointly emphasizes the central area of each image patch while keeping the contextual information. The central region features and contextual information are adaptively organized for the classification task.

\item Experimental results on three real SAR datasets demonstrate the effectiveness of the proposed DDNet and promising results are observed. To benefit other researches along remote sensing image change detection, the PyTorch code of the proposed DDNet has been made publicly available.

\end{itemize}

\section{Methodology}

Given two SAR images $I_1$ and $I_2$ captured at different times over the same geographical region, the objective is to generate a binary change map in which the changed pixels are marked as ``1" and unchanged pixels are marked as ``0".

The proposed model works in an unsupervised manner. The preclassification is the first step. The main purpose of this step is to find samples which have a high probability to be changed or unchanged. The log-ratio operator \cite{Bazi06_grsl} is first employed to generate a difference image (DI). Then, hierarchical FCM clustering \cite{gao16_grsl} is implemented to classify the DI into three clusters: $\Omega_c$, $\Omega_u$, and $\Omega_i$. Pixels belonging to $\Omega_c$ and $\Omega_u$ are reliable pixels which have high probability to be changed or unchanged, respectively. Pixels in $\Omega_i$ are uncertain and need to be further classified. 10\% of the image patches centered at pixels in $\Omega_c$  and $\Omega_u$ are randomly chosen as the training samples for DDNet. It should be noted that the number of positive samples and negative samples are equal. For a given pixel, image patches centered at the pixel are extracted from $I_1$ and $I_2$, respectively. The size of each patch is $r \times r$ ($r=7$ in this work). Both patches are combined to form a new image patch with the size of $2 \times r \times r$. The generated new patches are fed into the DDNet for training. After training, image patches centered at pixels in $\Omega_i$ will be classified by the network. The whole process is unsupervised.

\begin{figure}[h]
\centering
\includegraphics [width=3.5in]{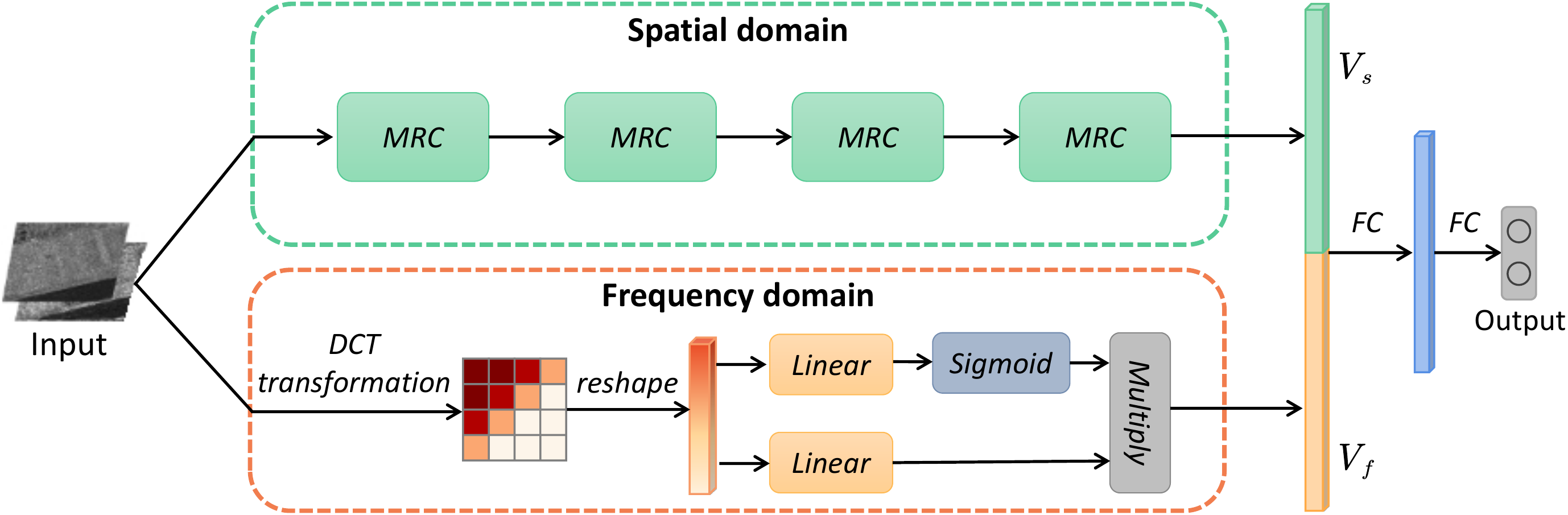}
\caption{An overview of the Dual-Domain Network (DDNet). The network is comprised of two branches: A spatial domain branch to capture multi-region features, and a frequency domain branch to encode the DCT coefficients. In the spatial domain branch, the network contains four MRC modules, which is able to emphasize the central region features while retaining the contextual information. In the frequency domain branch, the input image patch is converted to the frequency domain by DCT, and then an ``on-off switch" is employed to select critical components of the DCT coefficients.}
\label{fig_frame}
\end{figure}

The framework of the proposed DDNet is illustrated in Fig. \ref{fig_frame}. The network is comprised of a spatial domain branch and a frequency domain branch. The spatial domain branch is designed to capture multi-region features, while the frequency domain branch is used to encode the DCT coefficients. Next, we introduce the spatial and frequency domain feature extraction, respectively. Afterwards, we show how to combine the features from both domains for further classification.

\subsection{Spatial Domain Feature Extraction}

\begin{figure}[h]
\centering
\includegraphics [width=3.4in]{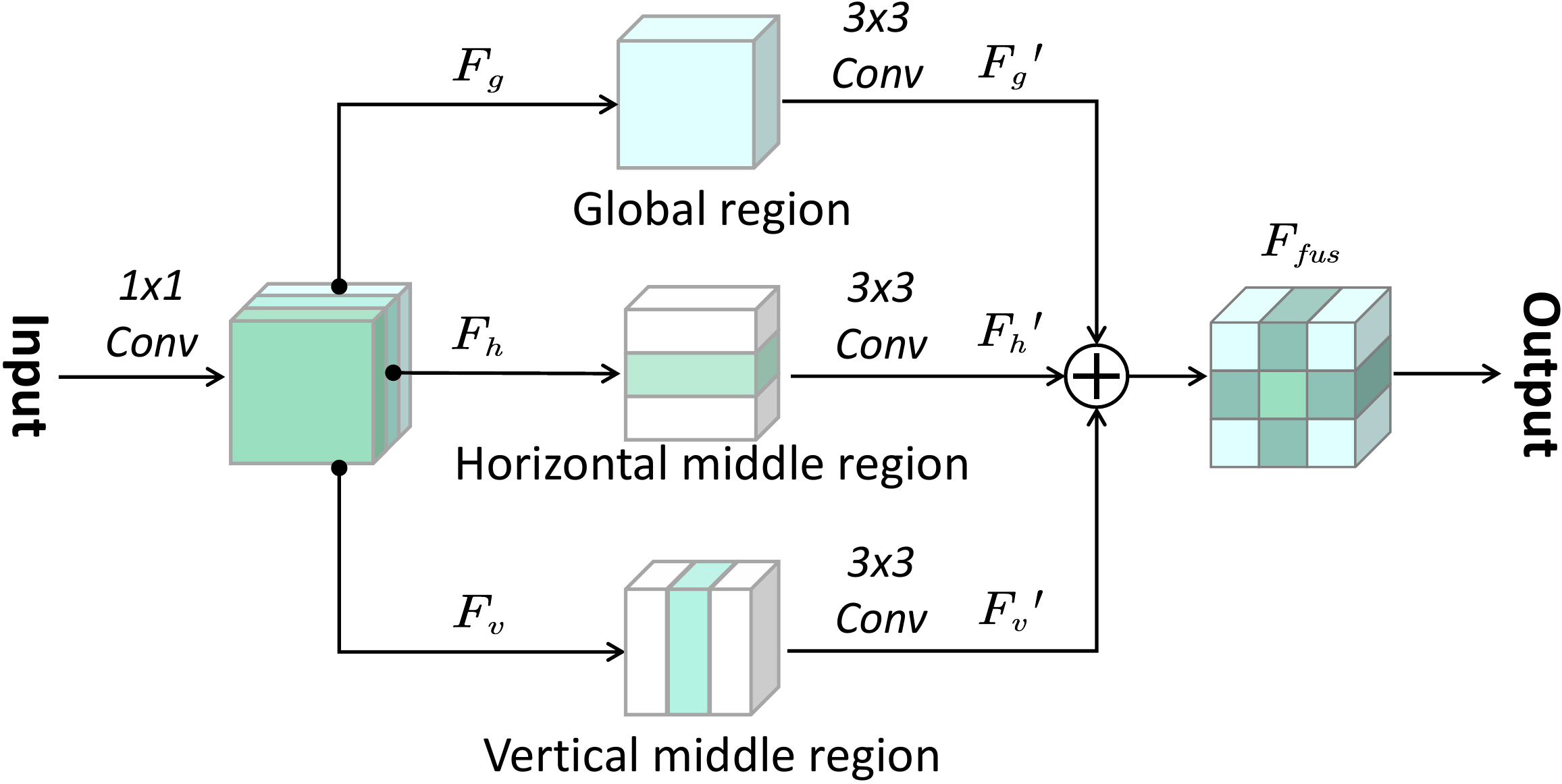}
\caption{Illustration of the MRC module. Taking patch size $r=7$ as an example. The input feature maps are convolved into 15 channels, and then divided into three groups $F_g$, $F_h$, and $F_v$. After the $3\times3$ convolution layer, three groups of features $F_g'$, $F_h'$, and $F_v'$ can be obtained. These features are fused with element-wise summation to form the output features. }
\label{fig_mrc} 
\end{figure}

In the spatial domain, the network contains 4 multi-region convolution (MRC) modules as illustrated in Fig.\ref{fig_frame}. Details of the MRC module are shown in Fig.\ref{fig_mrc}. Since the contextual information is essential for SAR image change detection, existing methods commonly adopt a windows with fixed size (3$\times$3, 5$\times$5, 7$\times$7, etc) to determine whether the position has changed. We argue that if we abandon some marginal regions in feature extraction, the central region can be emphasized and the noise in marginal regions can be potentially eliminated. Toward this end, we propose to extract features of multi-regions to enhance the feature representation in SAR change detection. 

The follows are three representative situations: 1) \textit{Global region}. As illustrated in Fig. \ref{fig_mrc}, it is a candidate of square-shaped patch, where the CNN is guided to capture the global contextual information of the central pixel. 2) \textit{Horizontal middle region}. The image patch is designed to concentrate on the center area, and pixels on the top and bottom have been removed. 3) \textit{Vertical middle region}. The image patch is designed to concentrate on the center area, and pixels on the left and right have been removed. If the above-mentioned regions are considered in CNN model, the central region will be emphasized and the noisy pixels on the margin will be effectively suppressed. 

Given an image patch $A\in \mathbb{R}^{2 \times r\times r}$, we feed it into a $1\times1$ convolution layer to generate new feature map $F\in \mathbb{R}^{C\times r\times r}$. Then, $F$ is divided into three groups $F_g$, $F_h$ and $F_v$ according to the channel dimension. Hence, the shape of $F_g$, $F_h$, and $F_v$ is $\frac{C}{3}\times r\times r$, respectively. $F_g$ represents the global region features. $F_h$ denotes the horizontal middle region feature, where several top and bottom rows are set to 0. $F_v$ denotes the vertical middle region feature, where several left and right columns are set to 0. It should be noted that $r$ is set to 7 in this work, two top rows and two bottom rows are set to 0 in $F_h$, while two left and two right columns are set to 0 in $F_v$.

After the $3\times3$ convolution layer, we obtain three groups of features $F_g'$, $F_h'$ and $F_v'$, respectively. These features are fused by element-wise summation as: 

\begin{equation}
  F_{fus} =  F_g' + F_h' + F_v',
\end{equation}
where $F_{fus}\in \mathbb{R}^{\frac{C}{3}\times r\times r}$ represents the spatial fused features. $C$ is set to be 15 in our implementations, and we therefore obtain the final spatial fused feature map with the size of $5\times 7\times 7$. The features are then reshaped to a vector $V_s$, and the vector is $5\times 7\times 7 = 245$ in length. Hence, $V_s$ has a global context view and the center region information has been enhanced.

\subsection{Frequency Domain Feature Extraction}

Due to the presence of speckle noise, it is difficult to extract robust discriminative features from the DI in the spatial domain. Recently, researchers have verified that the compressed representation in the frequency domain is capable of suppressing the speckle noise \cite{wu18_cvpr}. Inspired by Xu's work \cite{xu20_cvpr}, discrete cosine transform (DCT) is employed for frequency feature extraction. 

An input image patch of size $2\times r\times r$ is resized to $2\times 8 \times 8$ by bilinear interpolation. Then, the image patch is converted to the frequency domain by DCT. Afterwards, the obtained DCT coefficient vector $v$ is $2\times64 =128$ in length. It has been confirmed that DCT proves highly effective in noise suppression in the spatial domain. In order to further select the critical components of the DCT coefficients, we employ an ``on-off switch" which generates an informative vector $i$ and an attention gate $g$ with two linear transformations, as illustrated in Fig. \ref{fig_frame}. The informative vector $i$ and attention gate $g$ are generated as:

\begin{equation}
    i = W^iv+b^i,
\end{equation}
\begin{equation}
    g = \sigma(W^gv+b^g),
\end{equation}
where $W^i$ and $W^g$ are the weight matrix, $b^i$ and $b^g$ are the bias in linear transformation, and $\sigma$ denotes the sigmoid activation function.

Then, the DCTB employs the attention gate to the information vector using element-wise multiplication and obtains the final frequency feature vector $V_{f}$ as:
\begin{equation}
    V_{f} = g \odot i,
\end{equation}
where $\odot$ is the element-wise multiplication. 

\subsection{Final Change Map Generation}

After obtaining the spatial domain feature $V_s$ and frequency domain feature $V_f$, they are concatenated and fed into a fully connected (FC) layer. Then, the possibility of changed or unchanged is calculated by a softmax layer to generate the output. After training, pixels in $\Omega_i$ will be classified by the network, and the final change map can be obtained. 

\begin{figure*}[htb]
  \centering
  \includegraphics[width=6.5in]{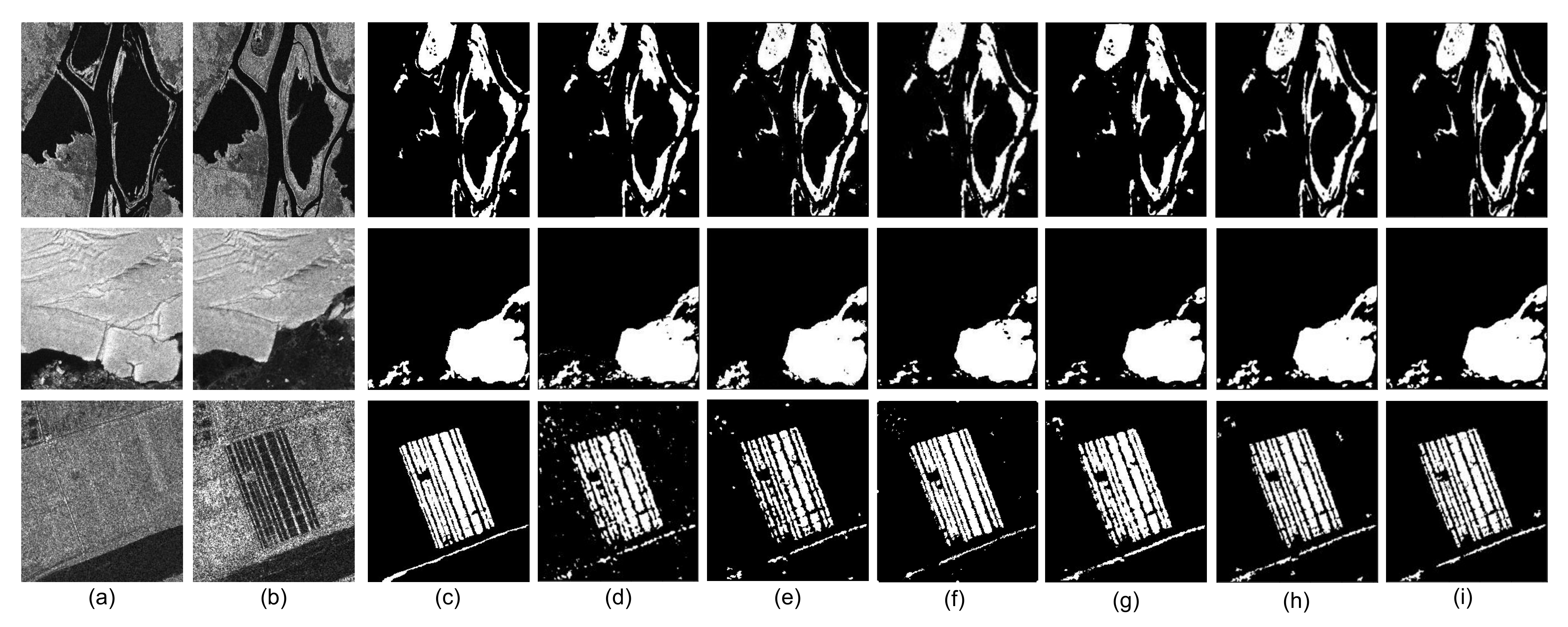}
  \caption{Visualized results of different change detection methods on the Ottawa dataset (first row), Sulzberger dataset (second row) and Yellow River dataset (third row) :  (a) Image captured at $t_1$. (b) Image captured at $t_2$. (c) Ground truth image.(d) Result by PCANet \cite{Celik09_grsl}. (e) Result by NR-ELM\cite{Gao16_jars}. (f) Result by DBN \cite{Gong16_trans}. (g) Result by DCNet \cite{Gao19}. (h) Result by MSAPNet \cite{Wang20_igarss}. (i) Result by the proposed DDNet.}
  \label{fig_result}
\end{figure*}

\section{Experimental Results and Analysis}

In this section, extensive experiments are conducted using three real SAR datasets. First, we describe the datasets used in the experiment and give the evaluation criteria. Then, the critical parameters that affect the change detection result are discussed. Finally, the DDNet is compared with several state-of-the-art methods.

\subsection{Dataset and Evaluation Metric}

To verify the effectiveness of the proposed DDNet, we use three co-registered and geometrically corrected multitemporal SAR datasets acquired by different sensors. The first dataset is the Ottawa dataset, which was captured by the RADARSAT SAR sensor in May and August 1997. The size of each image is $290\times350$ pixels. The second dataset is the Sulzberger dataset with the size of $256 \times 256$ pixels. It is a part of Sulzberger Ice Shelf which was provided by the European Space Agency's Envisat satellite. This image shows the breakup of an ice shelf caused by a tsunami in March 2011. The last dataset is the Yellow River dataset, which is from the Yellow River Estuary in China. It is acquired by the Radarsat-2 satellite in June 2008 and June 2009. The size of the dataset is $291 \times 306$ pixels. It is worth noting that it is difficult to identify the changed regions accurately since speckle noise is much stronger in the Yellow River dataset.

We measure the performance of the proposed DDNet with five common evaluation metrics in change detection, including false positives (FP), false negatives (FN), overall error (OE),  percentage of correct classification (PCC) and kappa coefficient (KC).

\subsection{Analysis of the Patch Size}

\begin{figure}
  \centering
  \includegraphics[width=2.8in]{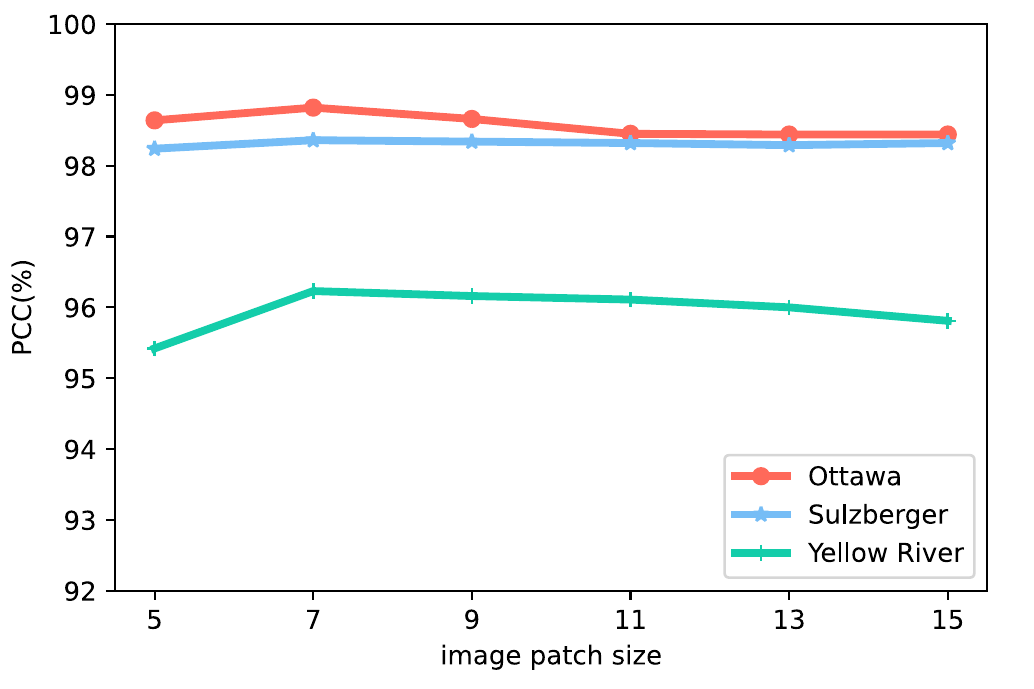}
  \caption{Relationship between different image patch size and the PCC.}
  \label{fig_patchsize}
\end{figure}

The spatial contextual information is captured by an image patch with the size of $r$. We evaluate the performance of the proposed DDNet by taking $ r=$ 5, 7, 9, 11, 13, and 15, respectively. Fig. \ref{fig_patchsize} shows that the relationship between $r$ and PCC. It can be observed that, as $r$ gradually increases, the value of PCC first rises and then tends to be stable. The PCC curves show that the spatial contextual information is significant for the change detection task. However, a large patch size increases the computational burden, and it may introduce some noise information that affects the change detection performance. Therefore, we take $r=7$ in our following implementation.

\begin{figure}
  \centering
  \includegraphics[width= 3in]{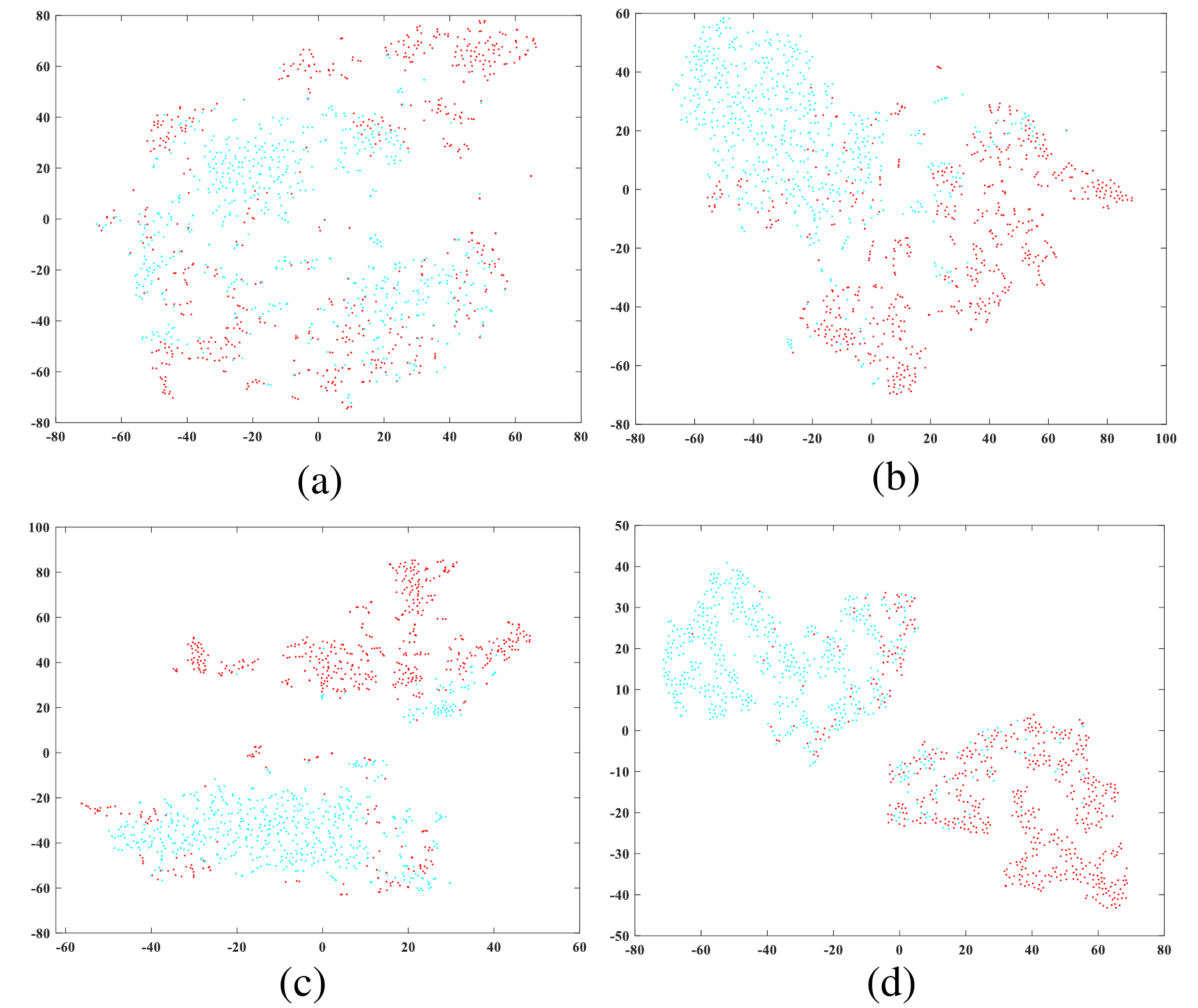}
  \caption{Visualization of the feature representations on the Ottawa dataset. (a) Input features. (b) Spatial domain features. (c) Frequency domain features. (d) Dual-domain features.}
  \label{fig_feature_vis}
\end{figure}

\subsection{Ablation Study}

In order to explore the effectiveness of MRC module and frequency domain features, ablation studies are conducted on three datasets. We designed three variants: 1) \textbf{CNN} denotes the traditional CNN, 2) \textbf{w/o DCT} refers to the model removing the frequency domain branch, and 3) \textbf{w/o MRC} denotes the model excluding the spatial domain branch. From Table. \ref{table_ablation}, it can be observed that removing either the DCT or MRC module affects the feature expressions and degrades the change detection performance. 

To further demonstrate the effectiveness of the MRC and DCT, we visualize the features of both domain in Fig. \ref{fig_feature_vis}. The representations learned in the dual domain are significantly better than the input with more separated and clearly bounded clusters.

\renewcommand\arraystretch{1.2}
\begin{table}[h]
\centering
\caption{Ablation Studies of the Proposed DDNet}
\label{table_ablation}
\begin{tabular}{l|c c c} \hline
Method & ~~Ottawa~~ & ~~Sulzberger~~ & Yellow River\\ \hline\hline
CNN       & 98.26 & 98.36  &95.60   \\
w/o DCT   & 98.34  & 98.63  & 95.91    \\  
w/o MRC  & 98.32  & 98.42  & 95.38    \\ 
Proposed DDNet     & 98.36  & 98.82  & 96.23    \\  \hline
\end{tabular}
\end{table}

\renewcommand\arraystretch{1.2}
\begin{table*}[htb]
    \centering
	\caption{Change Detection Results of Different Methods on Three Datasets.}
	\label{table_res}
    \begin{tabular}{c|c c c c c|c c c c c|c c c c c } \hline
    \multirow{2}{*}{Method} 
    & \multicolumn{5}{c|}{Results on the Ottawa dataset} & \multicolumn{5}{c|}{Results on the Sulzberger dataset} & \multicolumn{5}{c}{Results on the Yellow River dataset} \\ \cline{2-16}
      & FP & FN & OE & PCC  & KC & FP & FN & OE & PCC  & KC & FP & FN & OE & PCC  & KC\\ \hline \hline
     PCAKM \cite{Celik09_grsl}   & 955 & 1515 & 2470 & 97.57 & 90.73 & 711 & 429 & 1190 & 98.18 & 93.90  & 1829 & 2806 & 4635 & 93.76 & 78.32\\
     NR-ELM \cite{Gao16_jars}    & 695 & 1076 & 1771 & 98.26 & 93.38 & 719 & 832 & 1551 & 97.63 & 91.95  & 1059 & 3777 & 4836 & 93.49 & 76.14\\
     DBN \cite{Gong16_trans} & 995 & 704 & 1699 & 98.33 & 93.76  & 149 & 764 & 913 & 98.61 & 95.18  & 659  & 2674  & 3333  & 95.51 & 83.91 \\
     DCNet \cite{Gao19}  & 679 & 1051 & 1730 & 98.30 & 93.54  & 529 & 364 & 893 & 98.64 & 95.64  & 790 & 2137 & 2927 & 96.06 & 86.16 \\
     MSAPNet \cite{Wang20_igarss}  & 805 & 952 & 1757 & 98.27 & 93.47  & 503 & 396 & 899 & 98.63 & 95.60  & 817 & 2157 & 2974 & 96.00 & 85.94\\
     Proposed DDNet  & 641 & 1027 & 1668 & 98.36 & 93.77 & 264 & 511 & 775 & 98.82 & 96.17  & 952 & 1846 & 2798 & 96.23 & 86.95\\\hline
    \end{tabular}
\end{table*}

\subsection{Performance Comparison}

To verify the effectiveness of the proposed model, we compared DDNet with the following state-of-the-art baselines: PCAKM \cite{Celik09_grsl}, NR-ELM \cite{Gao16_jars}, DBN \cite{Gong16_trans} , DCNet \cite{Gao19} and MSAPNet \cite{Wang20_igarss}. It is worth mentioning that above methods are implemented with the default parameters described in their works. Table \ref{table_res} and Fig. \ref{fig_result} summarize the quantitative and visualized results of different methods, respectively.

On the Ottawa dataset (first row of Fig. \ref{fig_result}), we can observe that the result of PCAKM suffers from high FN value, which means that many changed regions are missed. The proposed DDNet achieves the best performance. The KC value of the proposed method is improved by 3.04\%, 0.39\%, 0.01\% , 0.23\% and 0.3\% over PCAKM, NR-ELM, DBN, DCNet and MSAPNet, respectively.  Meanwhile, on the Sulzberger dataset (second rows of Fig.\ref{fig_result}), it can be seen that the change maps generated by PCAKM and NR-ELM contain many noisy areas, and thereby the overall performance is affected. For NR-ELM, DBN and DCNet, the final change maps miss many changed regions, and they are afflicted with high FN values. Furthermore, deep learning-based methods have better performance than shallow model methods. The KC value of the proposed DDNet increased by 0.99\%, 0.53\% and 0.57\% over DBN, DCNet and MSAPNet on the Sulzberger dataset, respectively. From the above comparisons, it is evident that the proposed DDNet is capable of extracting features from the spatial and frequency domains in parallel effectively on the Ottawa and Sulzberger datasets.

The speckle noise on the Yellow River dataset is much stronger, and hence it is rather challenging to identify the changed regions accurately. On the Yellow River dataset (the third row of Fig. \ref{fig_result}), there are many small noisy areas in the final changed maps generated by PCAKM and NR-ELM. Therefore, the FN values of both methods are extremely high in Table \ref{table_res}. Compared with other methods, the proposed DDNet obtains 8.63\%, 10.18\%, 3.04\%, 0.79\% and 1.01\% gains in KC over PCAKM, NR-ELM, DBN, DCNet, and MSAPNet, respectively. 

Experimental results on the three datasets demonstrate that the proposed DDNet achieves the best performance, substantially surpassing all baselines. Besides, our model gains an advantage over traditional CNN-based models, which is mainly due to the joint learning of spatial and frequency domain features.

\section{Conclusion and Futuer Work}

In this letter, we present a novel DDNet to tackle the SAR change detection task. In the DDNet, the features of spatial and frequency domain are fused to improve the classification performance. In the spatial domain, we design a multi-region convolution module to enhance the central region features in the input image patches. In the frequency domain, we employ DCT and gating mechanism to extract the frequency features. Experimental results on three datasets demonstrated that the proposed DDNet is superior to several excellent change detection methods. In the future, we plan to work on large scale dataset to verify our change detection method.

\end{document}